\documentclass{article}
\usepackage{bbm}
\usepackage{float}
\usepackage{bm}
\usepackage{booktabs}
\usepackage{mathtools}
\usepackage{geometry}
\usepackage{hyperref}
\usepackage[capitalize]{cleveref}   
\usepackage{enumitem}
\usepackage{algorithm}
\usepackage{algpseudocode}
\usepackage{graphicx}
\usepackage{subcaption}
\usepackage{xparse}
\usepackage{amssymb}

\usepackage[most]{tcolorbox} 
\tcbuselibrary{skins}
\usepackage{xparse}
\usepackage{booktabs}
\usepackage{multirow}
\usepackage[dvipsnames]{xcolor}

\usepackage[
backend=biber,
style=alphabetic,
sorting=nty
]{biblatex}
\graphicspath{ {./images/} }
\hypersetup{colorlinks=true, linkcolor=blue, citecolor=blue, urlcolor=blue}

\title{\textbf{SNOO: Step-$K$ Nesterov Outer Optimizer \\ 
\Large The Surprising Effectiveness of Nesterov Momentum Applied to Pseudo-Gradients}}

\author{
\begin{tabular}{l@{\hspace{5em}}l@{\hspace{5em}}l}
\textbf{Dominik Kallusky\thanks{Equal contribution; alphabetical order}} & \textbf{Vinay Rao\footnotemark[1] \thanks{Work done while at Meta}} \\
Meta Platforms, Inc. &  \\
Menlo Park, CA & Palo Alto, CA \\
\texttt{dominik.kallusky@gmail.com} & \texttt{sr.vinay@gmail.com} \\
\\
\textbf{Vishal Nandavanam} & \textbf{Hao-Jun Michael Shi} \\
Meta Platforms, Inc. & Meta Platforms, Inc. \\
Menlo Park, CA & Menlo Park, CA \\
\texttt{vishal.nandavanam@gmail.com} & \texttt{hjmshi@meta.com} \\
\\
\end{tabular}
}

\usepackage{xspace}
\newcommand{\snoo}{SNOO\xspace}

\newcommand{\T}{\mathcal{T}}

\DeclareMathOperator{\AllReduce}{\texttt{AllReduce}}

\begin{document}
\date{}

\maketitle
\begin{abstract}
    The rapid development of large language models (LLMs) has driven the demand for more efficient optimization techniques. Among these, the Lookahead family of optimizers employs a two-loop framework, maintaining fast and slow sets of model weights. Multiple inner optimizer steps on the fast weights produce a trajectory -- the \textit{pseudo-gradient} -- that is used to update the slow weights. DiLoCo, a notable example originally designed for distributed training, applies Nesterov momentum to the averaged pseudo-gradient from multiple workers, claiming to even outperform AdamW in a non-distributed setup. In this paper, we empirically show that DiLoCo's surprising effectiveness stems primarily from applying Nesterov momentum to the pseudo-gradient, which improves training in a non-distributed setting. We call this Lookahead variant the Step-$K$ Nesterov Outer Optimizer (\snoo). We demonstrate that \snoo achieves compute factor gains of 1.5 -- 2.5$\times$ in a non-distributed setting up to a scale of 1e23 training FLOPs, with improvements that increase with model size. Because of its minimal compute and memory overhead and compatibility with model sharding, \snoo is a practical enhancement for a variety of inner optimizers, including AdamW and Muon.
\end{abstract}

\section{Introduction}
\label{sec:intro}

Scaling laws have profoundly influenced the pursuit of artificial general intelligence, motivating the development of increasingly large and capable language models \parencite{kaplan2020scalinglawsneurallanguage, hoffmann2022trainingcomputeoptimallargelanguage}.
As a result, pretraining costs for state-of-the-art models continue to escalate, driving the need for scalable, robust, and efficient training algorithms. For years, Adam \parencite{kingma2017adammethodstochasticoptimization} and its variant AdamW \parencite{loshchilov2017decoupled} have been the de facto optimizers for training large language models (LLMs). However, recent advances in second-order and adaptive methods, such as Shampoo \parencite{shampoo, shi2023distributed, anil2021scalablesecondorderoptimization}, SOAP \parencite{vyas2024soap,eschenhagen2025purifying}, AdEMAMix \parencite{pagliardini2024ademamix} and Muon \parencite{carlson2015preconditioned,tuddenham2022orthogonalising,jordan2024muon} have demonstrated significant improvements in convergence in terms of per-sample efficiency at scale \parencite{kimiteam2025kimik2openagentic}.

Another promising direction in optimization research for deep learning involves \textit{two-loop optimization strategies}. 
Methods such as the Lookahead optimizer \parencite{zhang2019lookaheadoptimizerksteps} maintain two sets of model weights -- ``fast'' and ``slow'' -- which are updated at different time scales. An inner optimizer performs several steps to update the fast weights, also known as the \textit{pseudo-gradient} direction, and the slow weights are periodically updated by interpolating towards the fast weights, with the fast weights then reset to the slow weights. This approach smooths the training trajectory by reducing the variance of the weight updates and improves convergence.

Simultaneously, federated learning and local SGD methods such as FedAvg~\cite{mcmahan2017communication}, SlowMo~\cite{wang2020slowmo}, FedMom~\cite{huo2020faster}, and DiLoCo~\cite{douillard2024dilocodistributedlowcommunicationtraining, charles2025communicationefficientlanguagemodeltraining} also employ a two-loop optimization structure, but are specifically designed for large-scale distributed training. These approaches aim to reduce communication overhead caused by frequent, network-intensive synchronization between workers~\cite{stich2019local} by allowing each worker's weights to evolve independently for multiple optimizer steps before \textit{averaging} the local model weights. However, communication-efficient methods like local SGD and FedAvg are known to be less token-efficient than their non-distributed alternatives, particularly when data is heterogeneous~\cite{li2020convergencefedavgnoniiddata}.

Unlike other local SGD methods, DiLoCo distinguishes itself by unexpectedly \textit{outperforming} the non-distributed baseline \cite{douillard2024dilocodistributedlowcommunicationtraining}. Rather than simply averaging the local weights across workers, DiLoCo computes each worker's pseudo-gradient based on the difference between the updated local (fast) model weights and an independent set of global (slow) weights. It then applies the Nesterov momentum step to the averaged pseudo-gradient to update the global model weights. This approach merges distributed averaging from local SGD with the pseudo-gradient mechanism in Lookahead, and further enhances it by applying Nesterov momentum to the pseudo-gradient when updating the global weights.

\begin{figure}[h!]
    \centering

    \begin{subfigure}[t]{0.48\textwidth}
        \centering
        \includegraphics[height=5cm]{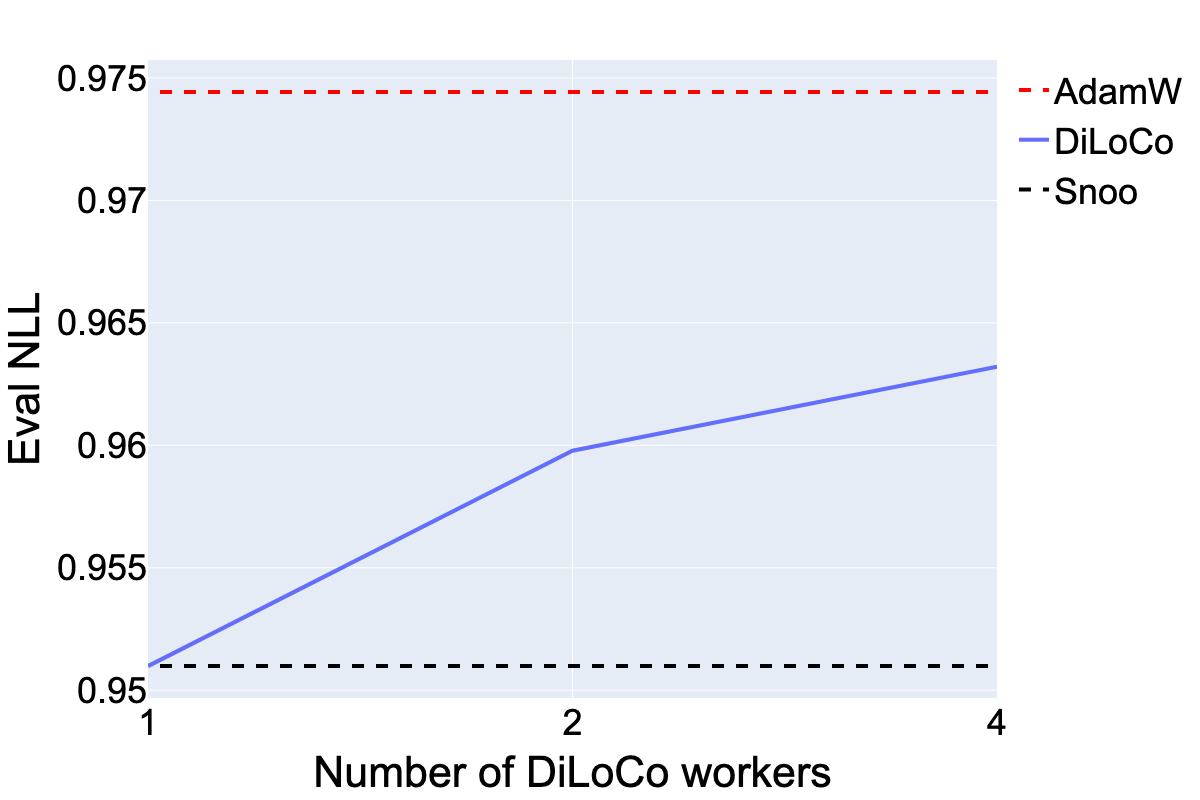}
        \caption{DiLoCo is able to outperform the AdamW baseline even when increasing the number of workers. However, its performance is best with only a single worker, which is equivalent to disabling the local SGD component and only applying Nesterov momentum to the pseudo-gradients.}
        \label{fig:worker_scaling1}
    \end{subfigure}
    \hfill
    \begin{subfigure}[t]{0.48\textwidth}
        \centering
        \includegraphics[height=5cm]{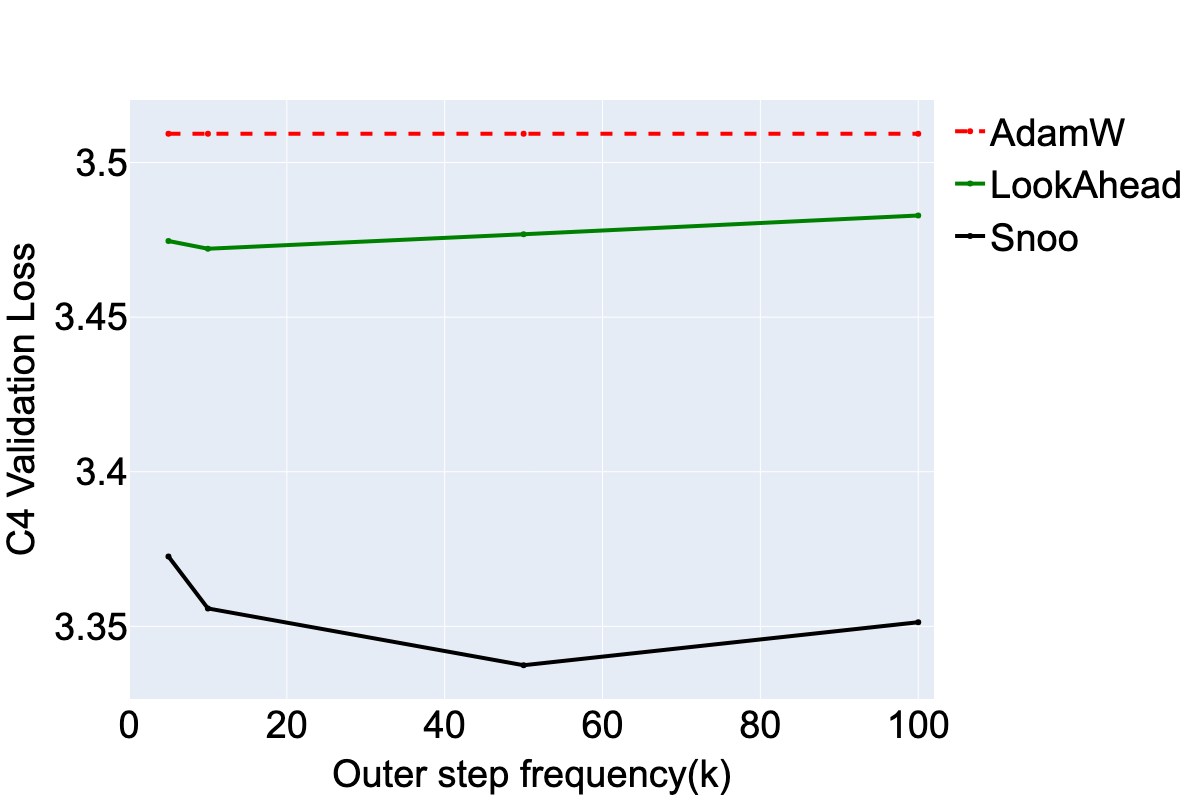}
        \caption{This figure plots the negative log-likelihood on the C4 validation set using TorchTitan on a 120M parameter Llama-3 model. \snoo outperforms the Lookahead (with AdamW) and AdamW optimizers, with an optimal choice of the outer step frequency $> 1$.}
        \label{fig:opt_comparison}
    \end{subfigure}

    \caption{Comparison of optimizer performance when varying the number of workers and number of inner steps.}
    \label{fig:combined_optimizers}
\end{figure}

In Figures \ref{fig:worker_scaling1} and \ref{fig:opt_comparison}, we empirically observe that DiLoCo’s strong performance can be primarily attributed to the Lookahead component with Nesterov momentum applied to the pseudo-gradient. 
Specifically, DiLoCo achieves its best performance when disabling local SGD by \textit{setting the number of workers to one}, but \textit{with a moderate number of inner optimizer steps}. This non-distributed version of DiLoCo demonstrates improved convergence over AdamW, motivating the study of the modified Lookahead component in DiLoCo alone, which we call the \textit{Step-$K$ Nesterov Outer Optimizer} (\snoo).

In this paper, we provide an in-depth scaling law study of \snoo across two settings: (1) small- and medium-scale open-source Llama-3 models trained on the C4 dataset; and (2) large-scale dense and MoE transformers with up to 1e23 FLOPs compute scale. Our experiments demonstrate that \snoo applied to the inner optimizer consistently outperforms the inner optimizer itself, with gains that become more pronounced as the model and data scale increase. Similar to exponential moving averaging of the iterates, we show that \snoo displays some intriguing implicit regularization properties, generating models with smaller weight norms and is more robust to overfitting from training on duplicated data. Lastly, we analyze both the systems implications for \snoo, showing that it is easy to integrate into existing training pipelines and incurs negligible overhead (i.e., a memory overhead of $2d$ and compute overhead of $\mathcal{O}(d/K)$, where $d$ is the number of parameters and $K$ is the number of inner steps) in large-scale settings. Our results suggest that applying \snoo to the inner optimizer yields more efficient and robust LLM training at minimal cost. 

\section{The Step-$K$ Nesterov Outer Optimizer (\snoo)}
\label{sec:methodology}

We model the LLM pre-training problem as expected risk minimization:
\begin{equation}
    \min_{w \in \mathbb{R}^d} F(w) = \mathbb{E}_{\xi \sim \mathcal{D}}\!\left[ f(w; \xi) \right]
\end{equation}
where $w \in \mathbb{R}^d$ are the model parameters and minibatch $\xi$ of size $B$ is drawn from a stationary data distribution $\mathcal{D}$.
At each step $t$, the inner optimizer proposes an update $-\eta_t\T_t$ based on the map corresponding to the inner optimizer's update rule $\T_t$ and inner learning rate $\eta_t > 0$:
\begin{equation}
    w_{t + 1} = w_t - \eta_t \T_t(f,w_t; \xi_t).
\end{equation}
Typically, the inner optimizer update $\T_t$ has the form $\T_t := H_t m_t$, where $m_t \approx \nabla F(w_t) \in \mathbb{R}^d$ is the gradient estimator and $H_t \in \mathbb{R}^{d \times d}$ is a symmetric positive semi-definite matrix. For example, if $H_t = I$ and $m_t = \nabla f(w_t; \xi_t)$, then we recover the stochastic gradient method. Other choices of $H_t$ and $m_t$ yield other common methods, such as AdamW, Shampoo, SOAP, or Muon.

We provide the complete pseudocode for the \snoo algorithm in Algorithm~\ref{alg:snoo}.

\begin{algorithm}[H]
    \caption{Step-$K$ Nesterov Outer Optimizer (\snoo)}
    \label{alg:snoo}
    \begin{algorithmic}
        \Require Initialization $w_0 \in \mathbb{R}^d$, inner optimizer update $\T_t$, inner learning rate schedule $\{\tilde{\eta}_{t, k}\}$, outer step frequency $K \geq 1$, outer learning rate $\eta > 0$, and outer momentum $\mu \geq 0$.
        \State Initialize momentum buffer $b_{-1} \gets 0 \in \mathbb{R}^d$.
        \For{$t = 0, 1, \ldots, T - 1$}
            \State Synchronize fast weights $\tilde{w}_{t, 0} \gets w_t$.
            \For{$k = 0, 1, ..., K - 1$}
                \State Sample minibatch $\xi_{t, k} \sim \mathcal{D}$.
                \State $\tilde{w}_{t, k + 1} \gets \tilde{w}_{t, k} - \tilde{\eta}_{t, k} \T_{t, k}(f, \tilde{w}_{t, k}; \xi_{t, k})$ \Comment{Inner optimizer update.}
            \EndFor
            \State $s_t \gets w_t - \tilde{w}_{t, K}$ \Comment{Compute pseudo-gradient.}
            \State $b_t \gets \mu b_{t - 1} + s_t$ \Comment{Update momentum buffer for Nesterov.}
            \State $w_{t + 1} \gets w_t - \eta (\mu b_t + s_t)$ \Comment{Update slow weights via Nesterov momentum.}
        \EndFor
        \State \Return slow weights $w_T$
    \end{algorithmic}
\end{algorithm}

Similar to the Lookahead optimizer, \snoo wraps an inner optimizer by maintaining two sets of model weights and applying the inner optimizer $\T_t$ to the fast model weights. The pseudo-gradient measures the trajectory given by the difference between the fast model weights $\tilde{w}_{t, K}$ from the slow model weights $w_t$:
\begin{equation*}
    s_t = w_t - \tilde{w}_{t, K}
\end{equation*}
However, unlike Lookahead, \snoo applies gradient descent with Nesterov momentum to the pseudo-gradient when updating the slow weights. Setting the momentum parameter $\mu = 0$ reduces \snoo to the Lookahead optimizer; see Appendix \ref{app:pseudocode}, Algorithm \ref{alg:lookahead}. Similarly, using a single local worker in DiLoCo recovers \snoo.

Although \snoo is presented in a two-loop form in Algorithm \ref{alg:snoo}, its practical usage is better understood through a single-loop formulation that combines both the inner and outer loops. Like DiLoCo, the algorithm does \textit{not} reset the inner optimizer's states or learning rate schedule between outer steps, instead using a learning rate schedule tuned across all inner iterations. This allows \snoo to be applied on top of an existing inner optimizer without significant hyperparameter re-tuning. The single-loop formulation is provided in Appendix~\ref{app:pseudocode}, Algorithm~\ref{alg:snoo-single-loop}. 

From a computational perspective, the algorithm only adds $\mathcal{O}(d)$ FLOPs every $K$ iterations, and requires storing two two additional buffers ($b$ and $\tilde{w}$) of the same dimension as the model parameters. This overhead is negligible and compatible with sharding schemes used in large language models, such as Fully-Sharded Data Parallelism (FSDP) \cite{zhao2023pytorch} and Tensor Parallelism (TP), as discussed in Section~\ref{sec:sys_analysis}.

The use of Nesterov momentum as an outer optimizer in deep learning has some precedent. For example, the NALA optimizer \cite{zuo2024nala} applies Sutskever's formulation of Nesterov momentum \cite{sutskever2013importance} as an outer optimizer to enhance the Lookahead optimizer. However, unlike NALA, \snoo leverages the practical implementation of SGD with Nesterov momentum as found in PyTorch and JAX. Momentum and Nesterov momentum has also been incorporated into various optimizers, but only for a single inner optimizer step. For instance, NAdam \cite{nadamw} and LaProp \cite{ziyin2021laprop} propose different ways to integrate momentum into AdamW, with LaProp being most similar to \snoo in that it applies standard momentum to a single inner step. Similar momentum variants have also been used in implementations of Shampoo \cite{anil2021scalablesecondorderoptimization, shi2023distributed}. 

Momentum was also shown to be equivalent to stochastic primal averaging \cite{defazio2020momentum}, which can be interpreted as a coupled version of exponential moving averaging of the iterates, where the gradient or search direction is evaluated at the averaged iterate. This connection may help explain some of the generalization properties of \snoo, which we explore further in Section~\ref{sec:regularization}. For example, it is known that EMA models are more robust to label noise \cite{morales-brotons2024exponential}.

\section{Scaling Laws}
\label{sec:experiments}

To demonstrate the viability of \snoo for training large-scale models, we compare its performance against an optimized AdamW baseline across multiple settings. In Section~\ref{subsec:compute_factors}, we describe our methodology for generating compute factor multipliers, which enables us to evaluate efficiency gains of different training algorithms as the model size and number of tokens increase. Section \ref{subsec:medium_scale} presents our results on a small- to medium-scale language model based on the Llama-3 architecture \parencite{grattafiori2024llama}, evaluated in a reproducible open-source setting on the C4 dataset. Finally, in Section~\ref{subsec:large_scale}, we extend this comparison to a production environment, demonstrating substantial compute factor gains for both dense and Mixture-of-Experts (MoE) models.

\subsection{Power Law and Compute Factors}
\label{subsec:compute_factors}

To rigorously evaluate \snoo's computational efficiency as models scale, we use a regression-based framework grounded in power law scaling that relates compute (measured in FLOPs) to the negative log-likelihood (NLL) scores on different benchmarks \cite{kaplan2020scalinglawsneurallanguage,hoffmann2022trainingcomputeoptimallargelanguage}. Since language models improve predictably with increased compute, FLOPs serve as a unified metric that captures both model size and token budget (which are expected to scale proportionally \cite{hoffmann2022trainingcomputeoptimallargelanguage}), enabling fair comparisons across architectures. Unlike model size or training steps alone, FLOPs reflect the actual computational resources used -- including the forward, backward, and optimizer steps across all tokens -- providing a true measure of efficiency. By focusing on FLOPs, we assess models based on computational efficiency rather than the optimizer's per-step performance. 

We use the power law relationship:
\begin{equation}
\mathcal{L}(C) = a \cdot C^{-b} + c
\label{eq:power_law_abc}
\end{equation}
where $\mathcal{L}(C)$ is loss at compute $C$, $a > 0$ is the scaling coefficient, $b > 0$ is the scaling exponent, and $c \geq 0$ represents the y-intercept. At large compute scales, the value of a 1\% absolute gain becomes increasingly significant. Therefore, we report improvements in terms of \textit{compute factors} that quantify the computational efficiency advantage of one optimizer's model over another by measuring the relative compute required to achieve equivalent performance levels. 

Given two regression models with fitted scaling relationships
\begin{align}
\mathcal{L}_{\text{baseline}}(C) &= a_1 \cdot C^{-b_1} + c_1 \\
\mathcal{L}_{\text{experimental}}(C) &= a_2 \cdot C^{-b_2} + c_2
\end{align}
and a compute budget $C_{\text{exp}}$, the loss achieved by the experimental model is:
\begin{equation}
\mathcal{L}_{\text{target}} = \mathcal{L}_{\text{experimental}}(C_\text{exp}) = a_2 \cdot C_{\text{exp}}^{-b_2} + c_2.
\end{equation}
Therefore, the amount of compute required for the baseline model to reach the same loss as the experimental model is:
\begin{equation}
C_{\text{baseline}} = \left( \frac{\mathcal{L}_{\text{target}} - c_1}{a_1} \right)^{-1/b_1}.
\end{equation}

Using the compute budgets, the \textit{compute factor} $\gamma$ represents how much additional compute is required for the baseline model to match the experimental model's target performance. Mathematically, it is given by the ratio:
\begin{equation}
\gamma = \frac{C_{\text{baseline}}}{C_{\text{exp}}} = \frac{1}{C_{\text{exp}}} \left( \frac{a_2 \cdot C_{\text{exp}}^{-b_2} + c_2 - c_1}{a_1} \right)^{-1/b_1}
\end{equation}
When $\gamma > 1$, the experimental model is more compute-efficient; likewise, if $\gamma < 1$, the baseline model is more compute-efficient. If $\gamma = 1$, the models have the same compute factors, and neither approach outperforms the other.

\subsection{Open-Source Language Model}
\label{subsec:medium_scale}

To establish the efficacy of \snoo in a reproducible setting, we first compare \snoo with AdamW against a competitive AdamW baseline on an open-source model. In this case, we utilize the Llama-3 architecture within the TorchTitan OSS training framework \parencite{liang2025torchtitan}\footnote{Code for running the OSS experiments can be found here: \url{https://github.com/vishal9-team/torchtitan-snoo}.}, executed on a cluster of 32 H100 GPUs. We first describe the experimental setup and baseline hyperparameter tuning, followed by a compute factor analysis of \snoo.

\subsubsection{Experimental Setup}

Our experiments are conducted on the English portion of the C4 dataset, with performance evaluated on a held-out validation set \parencite{raffel2020exploring}. We trained a suite of dense Llama-3 models ranging from 125M to 1B parameters, with architectural scaling parameters chosen generally in accordance with Chinchilla scaling principles \parencite{hoffmann2022trainingcomputeoptimallargelanguage}. Specific model and training parameters are detailed in \cref{tab:oss_model_architectures}.

All models were trained using Fully Sharded Data Parallel (FSDP) \parencite{zhao2023pytorch} with a global batch size of $B = 64$ sequences and a sequence length of $L = 8,192$. The training budget was set to achieve a token-to-parameter ratio of approximately 30:1. Gradient clipping with a maximum norm of $1.0$ was employed to ensure stability.

The AdamW optimizer \parencite{kingma2017adammethodstochasticoptimization,loshchilov2017decoupled} serves as our competitive baseline. To ensure its robustness, we performed a learning rate sweep from $10^{-5}$ to $3 \times 10^{-2}$ with a weight decay of $0.01$, a learning rate warmup with the number of warmup steps as specified in \cref{tab:oss_model_architectures}, and a linear decay schedule with a minimum learning rate factor of $0.1$. A learning rate of $3 \times 10^{-4}$ yielded near-optimal validation loss across all model scales and was thus fixed for all baseline comparisons. We further fixed $\beta_1 = 0.9$, $\beta_2 = 0.95$, and $\epsilon = 1e\!-\!8$ as default values.

\begin{table}[ht]
    \centering
    \caption{Architectural and training hyperparameters for the open-source Llama-3 models.}
    \label{tab:oss_model_architectures}
    \begin{tabular}{lcccc}
        \toprule
        \textbf{Parameter} & \textbf{125M} & \textbf{300M} & \textbf{500M} & \textbf{1B} \\
        \midrule
        Dimension ($d_{\text{model}}$) & 768   & 1,024 & 1,280 & 1,792 \\
        Layers ($n_{\text{layers}}$)    & 12    & 20    & 22    & 23    \\
        Heads ($n_{\text{heads}}$)      & 12    & 16    & 10    & 14    \\
        \midrule
        Steps                           & 9,600   & 19,200  & 31,200  & 57,600  \\
        Warmup Steps                    & 1,200   & 4,000   & 4,000   & 7,500   \\
        Tokens (billions)               & 5.03    & 10.07   & 16.36   & 30.20   \\
        \midrule
        Total FLOPs                     & $1.29 \times 10^{19}$ & $4.37 \times 10^{19}$ & $1.05 \times 10^{20}$ & $3.27 \times 10^{20}$ \\
        \bottomrule
    \end{tabular}
\end{table}

\subsubsection{Results}

\begin{figure}[H]
    \centering
    \includegraphics[width=6cm]{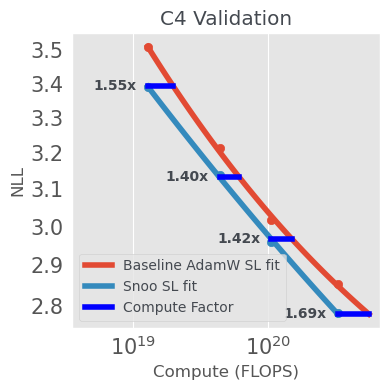}
    \caption{\snoo shows strong improvements across compute scales over the AdamW baseline on the C4 validation dataset. This figure plots NLL on a held-out validation set of C4 run using TorchTitan on OSS Llama-3 models.}
    \label{fig:llama3_scaling_law}
\end{figure}

The primary results of our experiments are summarized by the scaling laws for C4 validation loss (in terms of NLL), presented in Figure~\ref{fig:llama3_scaling_law}. For the \snoo{} optimizer, we performed a coarse grid search over its hyperparameters at each model scale: outer step frequency $k \in \{10, 50, 100\}$, outer learning rate $\eta \in \{0.5, 0.8, 0.95\}$, and outer momentum $\beta \in \{0.25, 0.5, 0.75\}$.

The compute factor analysis demonstrates that \snoo{} is consistently more compute-efficient than the AdamW baseline. This advantage, quantified using the compute factor metric (see \cref{subsec:compute_factors}), ranges from a $1.4\times$ speedup at smaller compute budgets to $1.69\times$ at the largest scale evaluated. While a power-law fit derived from a limited set of data points is susceptible to high variance and potential overfitting, the preliminary trend suggests \snoo's efficiency gains may become more pronounced at larger model scales.

\begin{figure}[H]
    \centering
    \includegraphics[width=\textwidth]{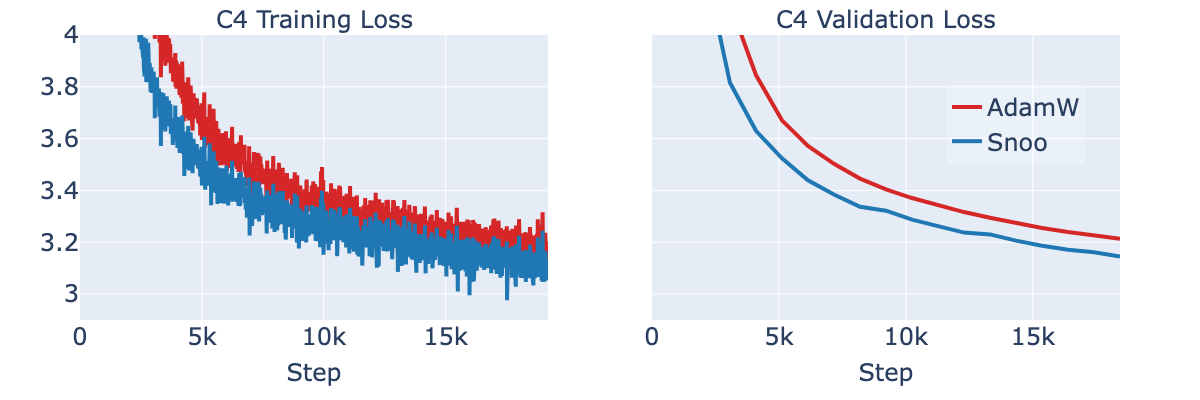}
    \caption{\snoo outperforms the AdamW baseline throughout training, exhibiting lower train and validation loss using a 300M dense transformer model. This figure plots NLL on training and held-out validation set of C4 run using TorchTitan on OSS Llama-3 models.}
    \label{fig:llama3_train_validation_loss}
\end{figure}

To provide intuition for this scaling advantage and to visualize the optimizers' behavior throughout training, we also present the training dynamics for an individual run. As illustrated in Figure~\ref{fig:llama3_train_validation_loss}, \snoo consistently achieves a lower validation loss than the baseline throughout training. For this run, the \snoo configuration utilized $K = 100$, $\eta = 0.8$, and $\mu = 0.75$. Critically, \snoo reaches the baseline's final validation loss in only 78\% of the training steps, corresponding to the $1.28\times$ training speedup in total steps.

\subsection{Large Scale Language Model}
\label{subsec:large_scale}

In order to validate our results at a larger scale, we use a real-world production setting for LLM pre-training and scale training up to 1e23 pre-training FLOPs. These experiments are trained on 2,048 H100 GPUs and use a pre-training dataset consisting of 25 trillion tokens.

\subsubsection{Experimental Setup}

We train both dense \cite{vaswani2017attention} and MoE \cite{shazeer2017outrageouslylargeneuralnetworks} transformer architectures. Hyperparameters such as learning rate, weight decay, and batch size are found through extensive sweeps, separately for both the inner optimizer and with \snoo; see Appendix~\ref{sec:hyperparams} for more details on the choice of optimal hyperparameters. The dense models are trained to a token-to-parameter ratio of 10:1, while the MoE models use a ratio of 45:1 (measured with \textit{active} parameter counts). The MoE has 32 experts and uses token-choice routing with $top_k=2$. All models are trained using the AdamW optimizer with a linear warmup and a cosine decay learning rate schedule. For evaluation, we showcase the following benchmarks: GPQA \cite{rein2023gpqagraduatelevelgoogleproofqa}, MMLU Pro \cite{wang2024mmluprorobustchallengingmultitask}, and Reasonbench \cite{zhang2025oedipussphinxbenchmarkingimproving}. Instead of the publicly available versions, we use slightly modified versions of these benchmarks.

\begin{figure}[H]
\centering
\includegraphics[width=1.0\textwidth]{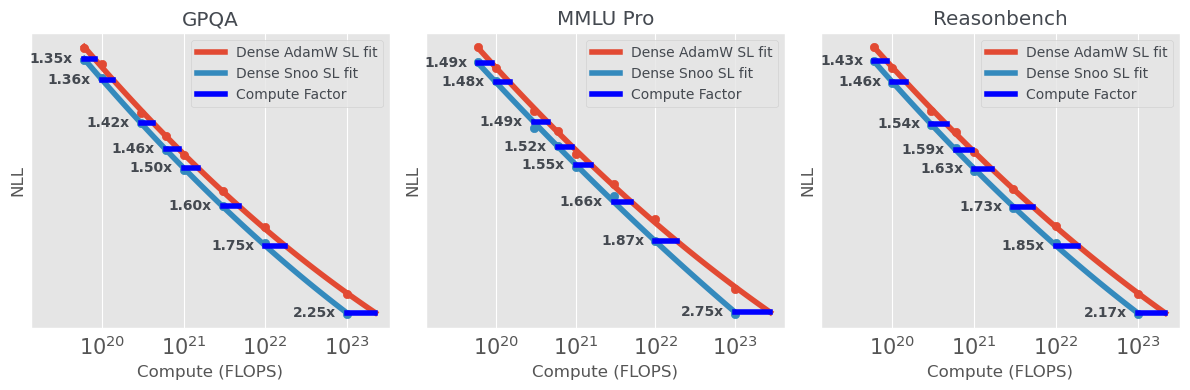}
\caption{\snoo outperforms the AdamW baseline across compute scales for dense models, with wins ranging from 1.35$\times$ through 2.75$\times$.}
\label{fig:dense_sl}
\end{figure}
\begin{figure}[h]
\centering
\includegraphics[width=1.0\textwidth]{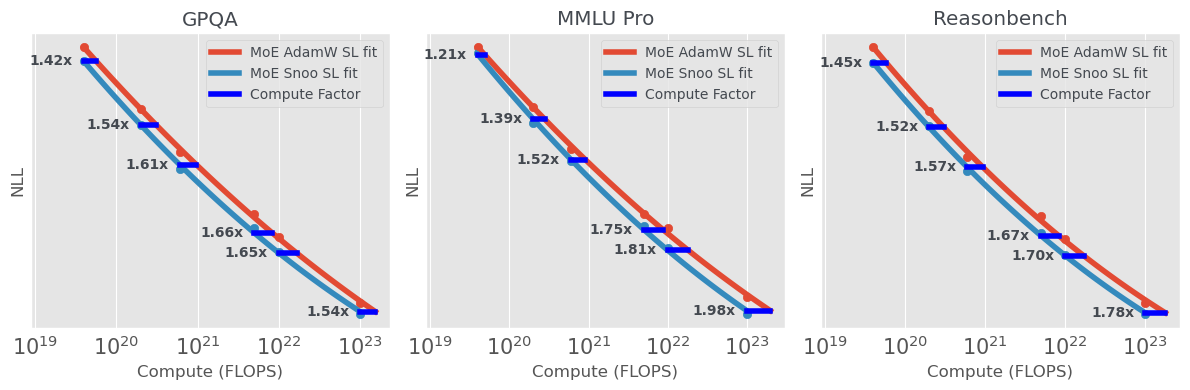}
\caption{\snoo outperforms the AdamW baseline across compute scales also for MoE models, with wins ranging from 1.2$\times$ through 2$\times$.}
\label{fig:moe_sl}
\end{figure}

\begin{table}[ht]
    \centering
    \label{tab:model_architectures}
    \small
    \begin{tabular}{l|ccc|ccc}
        \toprule
        & \multicolumn{3}{c|}{\textbf{Dense}} & \multicolumn{3}{c}{\textbf{Mixture-of-Experts}}\\
        \textbf{Optimizer} &  \textbf{GPQA} & \textbf{MMLU Pro} & \textbf{Reasonbench} & \textbf{GPQA} & \textbf{MMLU Pro} & \textbf{Reasonbench}\\
        \midrule
        AdamW & 0.4282 & 0.7961 & 0.3617 & 0.4017         & 0.7264   & 0.3432\\
        \snoo (AdamW)     & 0.4141      & 0.7644 & 0.3486   & 0.3942         & 0.7036  & 0.3344 \\
        \midrule
        \textbf{$\Delta$ (\%)}               & \textbf{3.29\%} & \textbf{3.98\%}   & \textbf{3.62\%}  & \textbf{1.87\%} & \textbf{3.14\%} & \textbf{2.56\%} \\
        \bottomrule
    \end{tabular}
    \caption{\snoo (with AdamW) and AdamW's NLL benchmark scores at the 1e23 compute scale for dense and MoE model architectures. Here, we show \snoo's percent improvement in NLL over AdamW.}
    \label{tab:large_scale_fits}
\end{table}

\subsubsection{Results}

Figures~\ref{fig:dense_sl} and~\ref{fig:moe_sl} show that compute factor gains range from $1.5\times$ to $2.5\times$, indicating that \snoo can deliver \textit{significant} improvements in optimization. Consistent with our open-source results, we observe that these gains increase with scale -- larger models and/or higher token budgets benefit more from \snoo than smaller ones. At the largest scales, \snoo can more than double training convergence speed, making it especially well-suited for large-scale LLM pre-training, where efficiency gains can quickly translate into substantial time and cost savings.

\section{\snoo's Implicit Regularization Properties}
\label{sec:regularization}

Beyond improved convergence, we hypothesize that one of the reasons for \snoo's improved performance is its implicit regularization effect, similar to exponential moving averaging of the iterates, which enables models trained by \snoo to generalize better than those trained by the inner optimizer alone. We present two empirical observations supporting this hypothesis.

\begin{figure}[htbp]
    \centering
    \begin{subfigure}[b]{0.45\textwidth}
        \includegraphics[width=\textwidth]{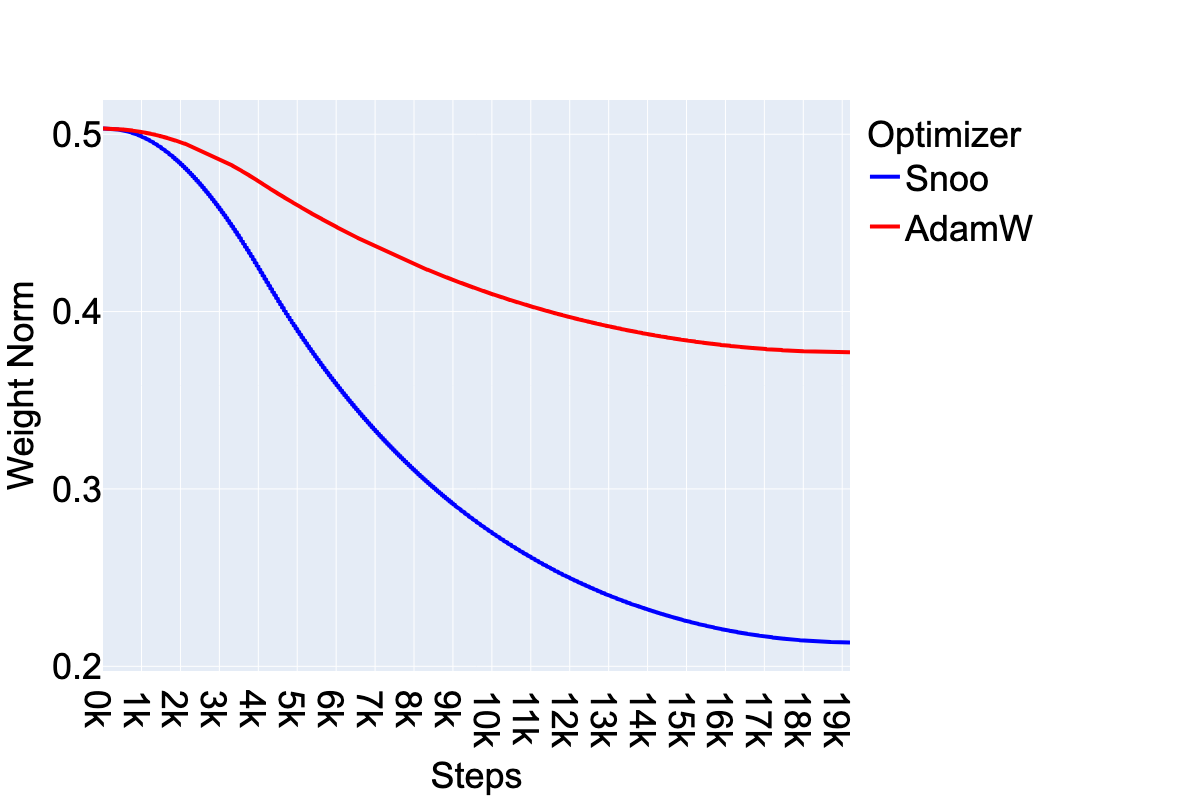}
        \caption{\snoo (with AdamW as the inner optimizer) encourages smaller $\ell_2$-norms of the model weights compared to AdamW, with weight norms continually decreasing as training progresses.}
        \label{fig:llama3_weight_norm}
    \end{subfigure}
    \hspace{1cm}
    \begin{subfigure}[b]{0.45\textwidth}
        \includegraphics[width=\textwidth]{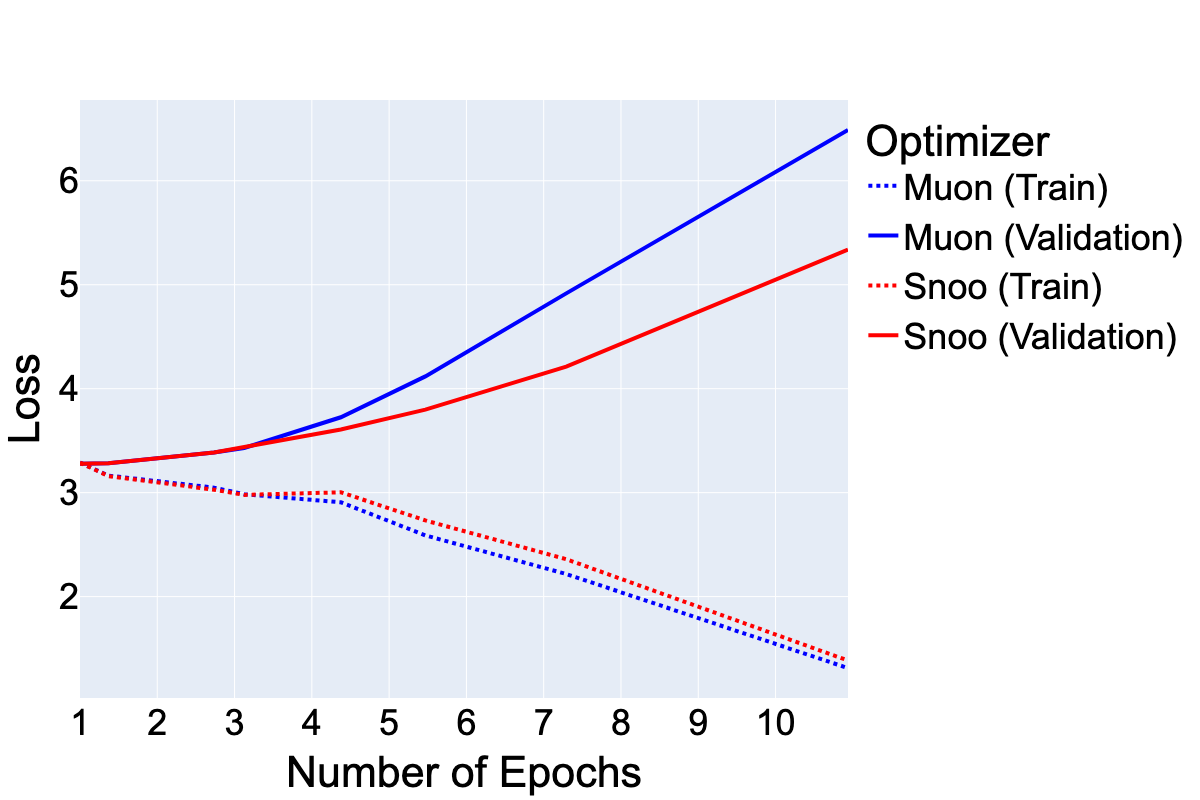}
        \caption{\snoo (with Muon as an inner optimizer) is more robust than Muon to overfitting. As the degree of data duplication increases, the gap between \snoo and Muon widens.}
        \label{fig:cont_nanogpt}
    \end{subfigure}
    \caption{\snoo exhibits better generalization properties compared to the AdamW and Muon baselines.}
\end{figure}

\textbf{Observation 1: \snoo introduces implicit weight regularization.}

We observe that \snoo regularizes model weights more effectively than the inner optimizer alone. To test this, we track the $\ell_2$-norm of the model weights throughout training for both the AdamW baseline and \snoo in the 300M model experiment from Section \ref{subsec:medium_scale}. As shown in Figure~\ref{fig:llama3_weight_norm}, models trained with \snoo consistently exhibit lower weight norms throughout the entire training progress, which continue to decrease as training progresses. We hypothesize that this effect may result from combining Nesterov momentum in the outer update with weight decay in the inner optimizer update, although this effect is not fully understood.
\\

\textbf{Observation 2: \snoo has better resilience to overfitting.}

Inspired by the regularization effects of exponential moving averaging, we further hypothesize that \snoo improves resilience to overfitting. To evaluate this, we used the \href{https://github.com/KellerJordan/modded-nanogpt}{modded-nanogpt} repository, a competitive speed-run framework based on \href{https://github.com/karpathy/nanoGPT}{nanogpt}. Using the hyperparameter settings from the current record-holding configuration that employs the Muon optimizer \cite{jordan2024muon}, we integrated \snoo as the outer optimizer. Notably, \snoo \textit{achieved the target loss in 5,640 iterations, surpassing the previous record of 5,690 iterations on the medium track}\footnote{See \url{https://github.com/KellerJordan/modded-nanogpt/pull/128}. Results continued to improve after our initial submission. While we also observed gains on the short track in terms of iteration count, these did not translate to faster wall-clock time. Further performance optimization is left for future work.}.

To test our hypothesis, we fixed the token and step budget to 1,750 steps (as used in the current record for the short track) and systematically duplicated the data by training on a subset of the dataset for multiple epochs. A single epoch corresponds to no repetition, while an epoch factor of 10 results in 10 identical epochs of 175 batches each. As seen in Figure~\ref{fig:cont_nanogpt}, increasing data duplication degrades the validation performance of both \snoo and Muon, even as training loss consistently improves for both \snoo and Muon. However, \snoo's validation loss degrades more slowly than Muon as the number of epochs increases, indicating greater robustness to overfitting. 

\section{Implementation and Systems Implications}
\label{sec:sys_analysis}

As discussed in Section~\ref{sec:methodology}, \snoo{} requires two additional buffers ($b$ and $\tilde{w}$) of size $d$, resulting in a total memory overhead of $2d$ parameters. However, in LLM pre-training, since the dominant contributor to memory usage is activation storage during the forward and backward passes, \snoo's $2d$ memory overhead is often negligible. For example, in transformer architectures with sequence length $L$ and batch size $B$, activation memory scales as $O(BdL)$, which can exceed parameter memory by an order of magnitude. Furthermore, the optimizer states can be sharded consistently with the model parameter's sharding through various parallelisms (FSDP and TP) without significant implementation challenges.

Because the outer updates occur only every $K$ steps, the infrequent access pattern of the buffers allows for efficient CPU offloading and asynchronous memory management. This minimizes contention for high-bandwidth GPU memory. Empirically, for $K \in [20, 100]$, the overhead associated with host-to-device (H2D) and device-to-host (D2H) transfers is amortized over $K$ steps and does not measurably impact training throughput. 

Regarding compute cost, since the Nesterov outer update is performed once every $K$ steps (where $K \in [20, 100]$ typically in practice), the compute overhead is less than $1\%$ of the total training step time for models of size $d \in [10^8, 10^{11}]$. This finding aligns with prior work on model averaging techniques \cite{zhang2019lookaheadoptimizerksteps, Xie2022AdanAN}, which report similarly negligible amortized costs.

\section{Conclusion}

In this paper, we investigated the surprising effectiveness of applying Nesterov-style momentum to the pseudo-gradients, unifying concepts from Lookahead \cite{zhang2019lookaheadoptimizerksteps} and DiLoCo \cite{douillard2024dilocodistributedlowcommunicationtraining}. Through extensive experiments spanning open-source and large-scale LLM pre-training (up to 1e23 training FLOPs), we found that the key driver of model quality improvements in algorithms like DiLoCo is the intermittent application of Nesterov momentum to the pseudo-gradients. This technique, which we call \snoo, can produce 1.5 -- 2.5$\times$ compute factor gains independent from cross-datacenter training.

Our empirical results suggest that \snoo not only improves convergence, but enhances the training algorithm's generalization properties, producing models with smaller weight norms and demonstrating stronger resilience to overfitting. However, the mechanisms behind these effects are not yet fully understood. Nesterov momentum's connection to iterate averaging \cite{defazio2020momentum} suggest that existing convergence and generalization theory for exponential moving averaging \cite{morales-brotons2024exponential} and primal averaging \cite{defazio2020momentum} may offer valuable insights into \snoo's effectiveness. Additionally, recent advances in analyzing DiLoCo \cite{khaled2025understandingouteroptimizerslocal} could be adapted to the single-worker case, providing another promising avenue for analyzing \snoo's convergence properties. 

\subsection*{Acknowledgements}
The authors would like to thank the Distributed Optimization Team at Meta where this work was incubated: Jiaming Cui, Aaron Defazio, Yuchen Hao, Dzmitry Huba, Vladimir Ivanov, Konstantin Mishchenko, Shangfu Peng, Parameswaran Raman, and Lin Xiao. We thank Parameswaran Raman for his comments on an initial version of the manuscript. We also thank Gregory Chanan, Davide Italiano, Sharan Narang, Maxim Naumov, Sandeep Parab, Joel Pobar, Prachi Sonalkar, and Chunqiang Tang for their managerial support. We thank Aurko Roy and Rohan Anil for helpful discussions.


\printbibliography

\newpage

\appendix
\section{Algorithm Pseudocode}
\label{app:pseudocode}

For completeness, we provide pseudocode for both the Lookahead optimizer and DiLoCo in Algorithms~\ref{alg:lookahead} and \ref{alg:diloco}. We provide the pseudocode for DiLoCo for each worker $i \in \{0, ..., W - 1\}$, where $W$ is the number of total local workers. We write them both in their original form. 

We also give the practical single-loop formulation of \snoo in Algorithm~\ref{alg:snoo-single-loop}. Note that both Lookahead and DiLoCo can be re-written in a similar formulation by combining the iterate counters $t$ and $k$ and periodically updating the slow weights $w$ every $K$ steps (specifically when $(t + 1) \bmod K = 0$).

\begin{algorithm}[H]
\centering
    \begin{algorithmic}
    \caption{Lookahead Optimizer}
    \label{alg:lookahead}
        \Require Initialization $w_0 \in \mathbb{R}^d$, inner optimizer update $\T_t$, inner learning rate schedule $\{\tilde{\eta}_{t, k}\}$, outer step frequency $K \geq 1$, and outer learning rate $\eta > 0$.
        \For{$t=0, 1, \ldots, T - 1$}
            \State Synchronize fast weights $\tilde{w}_{t, 0} \gets w_t$.
            \For{$k = 0, ..., K - 1$}
                \State Sample minibatch $\xi_{t, k} \sim \mathcal{D}$.
                \State $\tilde{w}_{t, k + 1} \gets \tilde{w}_{t, k} - \tilde{\eta}_{t, k} \T_{t, k}(f, \tilde{w}_{t, k}; \xi_{t, k})$ \Comment{Inner optimizer update.}
            \EndFor
            \State $s_t \gets w_t - \tilde{w}_{t, K}$ \Comment{Compute pseudo-gradient.}
            \State $w_{t + 1} \gets w_t - \eta s_t$ \Comment{Update slow weights.}
        \EndFor
        \State \Return slow weights $w_T$
    \end{algorithmic}
\end{algorithm}

\begin{algorithm}[H]
\centering
    \caption{DiLoCo (on Worker $i$)}
    \label{alg:diloco}
    \begin{algorithmic}
        \Require Initialization $w_0 \in \mathbb{R}^d$, number of local workers $W$, inner optimizer update for each worker $\{\T_{t, k}^{(i)}\}_{i = 1}^W$, inner learning rate schedule $\{\tilde{\eta}_{t, k}\}$, outer step frequency $K$, outer learning rate $\eta$, and outer momentum $\mu$.
        \State Initialize momentum buffer $b_{-1} \gets 0 \in \mathbb{R}^d$.
        \For{$t=0, 1, \ldots, T - 1$}
            \State Synchronize local (fast) weights $\tilde{w}_{t, 0}^{(i)} \gets w_t$.
            \For{$k = 0, ..., K - 1$}
                \State Sample minibatch $\xi_{t, k}^{(i)} \sim \mathcal{D}$.
                \State $\tilde{w}_{t, k + 1}^{(i)} \gets \tilde{w}_{t, k}^{(i)} - \tilde{\eta}_{t, k} \T_{t, k}^{(i)}(f, \tilde{w}_{t, k}^{(i)}; \xi_{t, k}^{(i)})$ \Comment{Inner optimizer update on worker $i$.}
            \EndFor
            \State $s_t^{(i)} \gets w_t - \tilde{w}_{t, K}^{(i)}$ \Comment{Compute local pseudo-gradient on worker $i$.}
            \State $s_t \gets \AllReduce(s_t^{(i)}) = \frac{1}{W} \sum_{i = 1}^W s_t^{(i)}$ \Comment{Synchronize pseudo-gradients across workers $i = 1, \ldots, l$.}
            \State $b_t \gets \mu b_{t - 1} + s_t$ \Comment{Update momentum buffer for Nesterov.}
            \State $w_{t + 1} \gets w_t - \eta (\mu b_t + s_t)$ \Comment{Update slow weights via Nesterov momentum.}
        \EndFor
        \State \Return global (slow) weights $w_T$
    \end{algorithmic}
\end{algorithm}

\begin{algorithm}[H]
    \caption{\snoo (Practical Single-Loop Formulation)}
    \label{alg:snoo-single-loop}
    \begin{algorithmic}
        \Require initialization $w_0 \in \mathbb{R}^d$, inner optimizer update $\T_t$, inner learning rate schedule $\{\tilde{\eta}_t\}$, outer step frequency $K \geq 1$, outer learning rate $\eta > 0$, and outer momentum $\mu \geq 0$.
        \State Initialize momentum buffer $b_{-1} \gets 0 \in \mathbb{R}^d$.
        \State Initialize fast weights $\tilde{w}_0 \gets w_0$.
        \For{$t=0, 1, \ldots, T - 1$}
            \State Sample minibatch $\xi_t \sim \mathcal{D}$.
            \State $\tilde{w}_{t + 1} \gets \tilde{w}_t - \tilde{\eta}_t \T_t(f, \tilde{w}_t; \xi_t)$ \Comment{Inner optimizer update.}
            \If{$(t + 1) \bmod K = 0$}
                \State $s_t \gets w_t - \tilde{w}_t$ \Comment{Compute pseudo-gradient.}
                \State $b_t \gets \mu b_{t - 1} + s_t$ \Comment{Update momentum buffer for Nesterov.}
                \State $w_{t + 1} \gets w_t - \eta (\mu b_t + s_t)$ \Comment{Update slow weights via Nesterov momentum.}
                \State $\tilde{w}_{t + 1} \gets w_{t + 1}$ \Comment{Synchronize fast weights with slow weights.}
            \Else
                \State $w_{t + 1} \gets w_t$ 
                \State $b_t \gets b_{t - 1}$ \Comment{Keep the momentum buffer and global weights the same.}
            \EndIf
        \EndFor
        \State \Return slow weights $w_T$
    \end{algorithmic}
\end{algorithm}

\section{Additional Tables}

\begin{table}[H]
\centering
\begin{tabular}{lllcccc}
\toprule
\textbf{Benchmark} & \textbf{Model} & \textbf{Optimizer} & \textbf{a} & \textbf{b} & \textbf{c} & \textbf{NLL\_fit @ 1e23} \\
\midrule
\multirow{4}{*}{GPQA Pro} & \multirow{2}{*}{MoE} & AdamW & 98.07 & -0.1213 & 0.2426 & 0.4017 \\
& & \snoo & 178.73 & -0.1369 & 0.2673 & 0.3942 \\
\cmidrule(lr){2-7}
& \multirow{2}{*}{Dense} & AdamW & 94.15 & -0.1214 & 0.2763 & 0.4282 \\
& & \snoo & 77.37 & -0.1164 & 0.2514 & 0.4141 \\
\midrule
\multirow{4}{*}{MMLU Alt} & \multirow{2}{*}{MoE} & AdamW & 185.32 & -0.1222 & 0.4398 & 0.7264 \\
& & \snoo & 352.62 & -0.1377 & 0.4636 & 0.7036 \\
\cmidrule(lr){2-7}
& \multirow{2}{*}{Dense} & AdamW & 88.97 & -0.1065 & 0.4801 & 0.7961 \\
& & \snoo & 41.37 & -0.0863 & 0.3360 & 0.7641 \\
\midrule
\multirow{4}{*}{Reasonbench} & \multirow{2}{*}{MoE} & AdamW & 208.53 & -0.1241 & 0.2308 & 0.3432 \\
& & \snoo & 250.01 & -0.1470 & 0.2326 & 0.3344 \\
\cmidrule(lr){2-7}
& \multirow{2}{*}{Dense} & AdamW & 58.99 & -0.1125 & 0.2092 & 0.3617 \\
& & \snoo & 62.89 & -0.1143 & 0.2008 & 0.3486 \\
\bottomrule
\end{tabular}
\caption{AdamW and \snoo's scaling law coefficients used in Table \ref{tab:large_scale_fits}.}
\label{tab:benchmark_performance}
\end{table}

\section{Hyperparameter Tuning}
\label{sec:hyperparams}

\begin{figure}[H]
    \centering
    \begin{subfigure}[t]{0.48\textwidth}
        \centering
        \includegraphics[width=\textwidth]{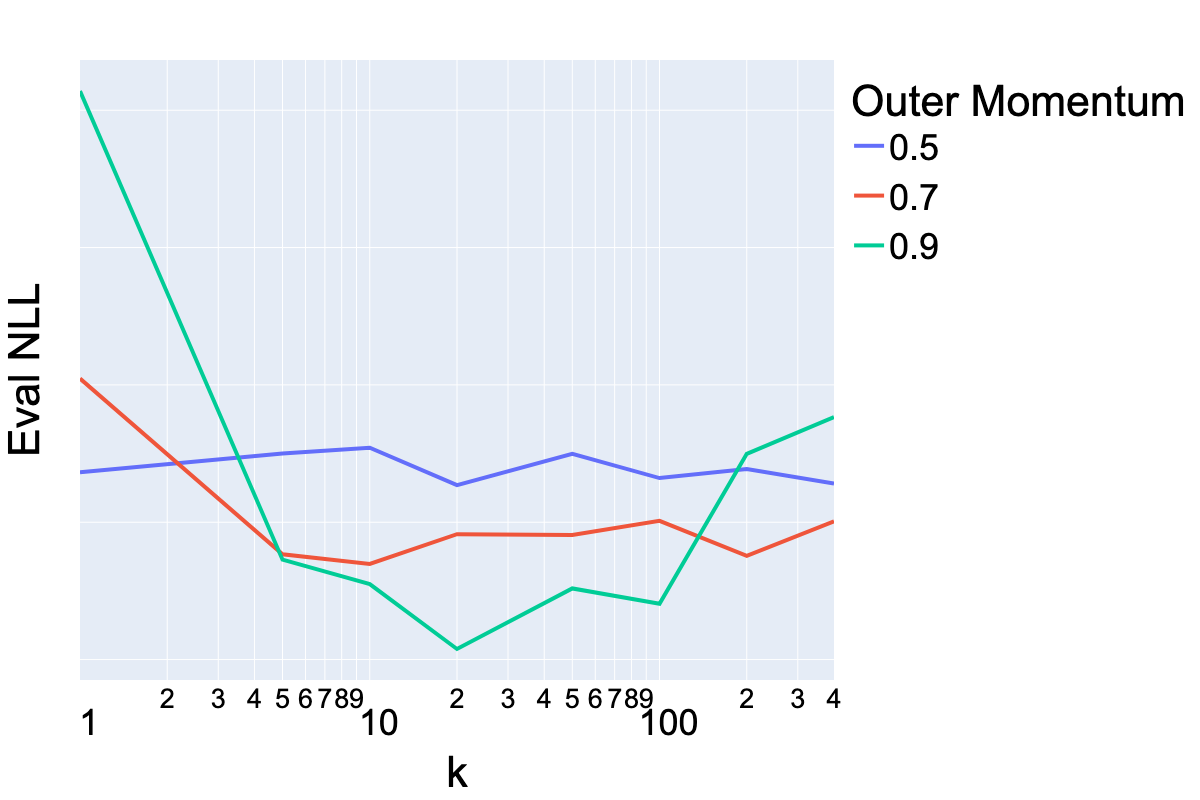}
        \caption{Ablation of the outer step frequency ($K$) for different values of the outer momentum coefficient ($\mu$). Values of $K \in [10, 100]$ perform best.}
        \label{fig:k_ablation_sub}
    \end{subfigure}
    \hfill
    \begin{subfigure}[t]{0.48\textwidth}
        \centering
        \includegraphics[width=\textwidth]{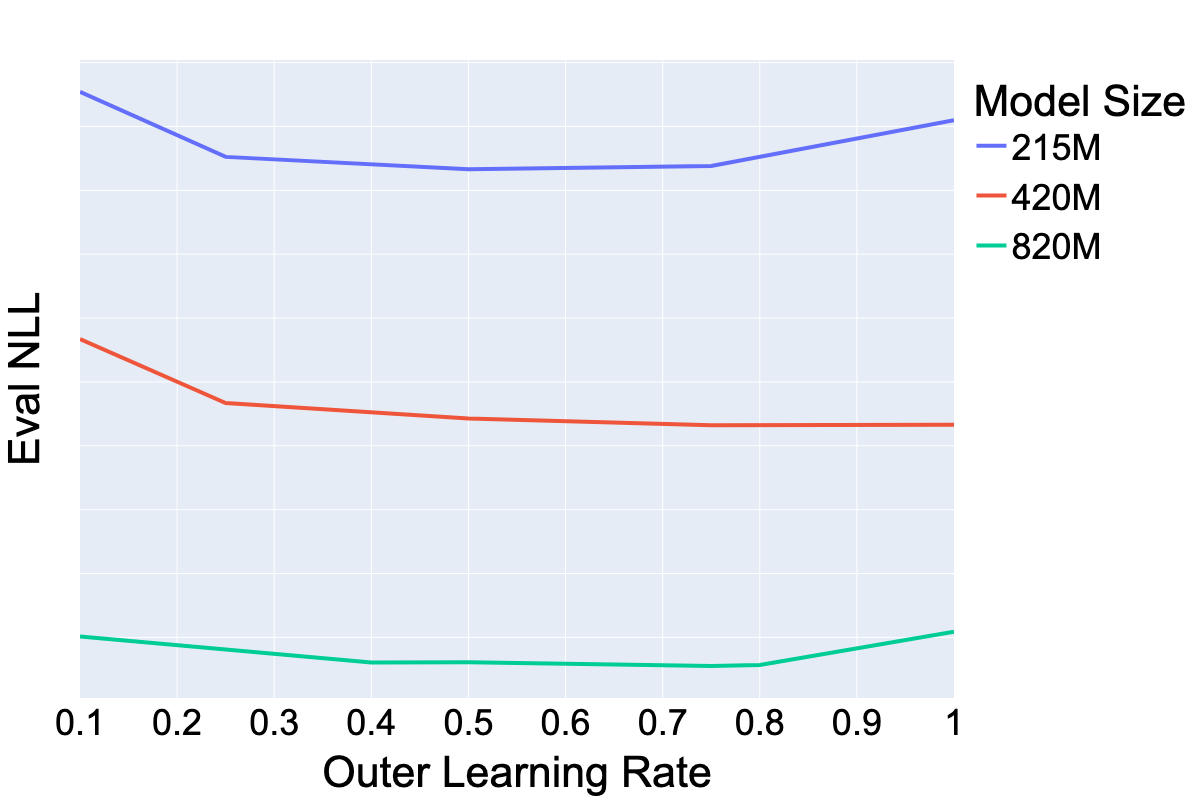}
        \caption{Ablation of the outer learning rate ($\eta$) across different model scales. \snoo performs best on a wide range of outer learning rates ($\eta \in [0.2, 0.8]$).}
        \label{fig:outer_lr_ablation_sub}
    \end{subfigure}
    \caption{Ablation studies on $K$, $\mu$, and $\eta$ shows that extreme values for both $K$ and $\eta$ are suboptimal.}
    \label{fig:ablation_combined}
\end{figure}

\snoo introduces three principal tunable hyperparameters: 
\begin{itemize}
    \item $\eta$: the outer learning rate;
    \item $\mu$: the outer momentum coefficient;
    \item $K$: the outer Nesterov step frequency.
\end{itemize}
These hyperparameters exhibit significant interactions, which prevents us from ablating each in isolation. Consequently, optimal performance requires joint tuning of $\eta$, $\mu$, and $K$. Our experiments indicate that the optimal configuration of these hyperparameters is further influenced by factors such as data mixture, model architecture (e.g., Mixture-of-Experts (MoE) versus dense models), the choice of inner optimizer, weight decay, and the inner learning rate. This interdependence complicates the hyperparameter search process, yet it is essential, as suboptimal choices can result in substantial performance degradation. In the subsequent analysis in Figure~\ref{fig:ablation_combined}, we adopt an experimental setup analogous to that described in Section~\ref{subsec:medium_scale}, and systematically ablate $\eta$, $\mu$, and $K$.

\subsection{Outer Step Frequency and Momentum}

We investigate the effect of varying the outer step frequency $K$ (ranging from 1 to 400) for outer momentum coefficients $\mu \in \{0.5, 0.7, 0.9\}$, using a fixed model architecture (Dense Transformer 210M). Our results in Figure~\ref{fig:k_ablation_sub} indicate that extreme values of $K$ are generally suboptimal. The highest performance is observed at $K = 20$, although a broad range ($K=5$ to $K=100$) yields comparably strong results. This finding suggests that Nesterov momentum is most effective when applied intermittently rather than after every step, unlike approaches like NAdamW \cite{nadamw} or LaProp \cite{ziyin2021laprop}.

We also observe a strong dependency between $K$ and $\mu$. Specifically, the momentum coefficient of $\mu = 0.9$ exhibits increased sensitivity to the outer step frequency, but achieves the highest performance when paired with the optimal choice of $K$. In contrast, a lower momentum coefficient ($\mu=0.5$) demonstrates relative invariance to the outer step frequency, albeit with consistently suboptimal results.

\subsection{Outer Learning Rate}

We also ablate the outer learning rate $\eta$ for three Dense Transformer models of sizes $\{215\text{M}, 420\text{M}, 820\text{M}\}$. In Figure \ref{fig:outer_lr_ablation_sub}, we observe a U-shaped performance curve across all model scales, indicating that extreme values of $\eta$ are suboptimal. For each model scale, there exists a broad region of near-optimal choices for $\eta$ in the range of $0.2$ -- $0.8$. Notably, the optimal value of $\eta$ remains relatively stable across all model sizes.

\end{document}